\documentclass[letterpaper, 10 pt, conference]{ieeeconf}  

\IEEEoverridecommandlockouts                              

\usepackage{amsmath} 
\usepackage{amssymb}  

\usepackage{algorithm}
\usepackage{algpseudocode}

\usepackage{graphicx, float, listings, hyperref}
\usepackage{cite}
\usepackage{amsmath, amssymb, bm}

\usepackage{multirow}
\usepackage{caption}
 
\usepackage{mathtools}
\usepackage{siunitx}
\usepackage{soul}
 
\usepackage{optidef}

\title{\LARGE \bf
Leveraging Symbolic Algebra Systems to Simulate Contact Dynamics in Rigid Body Systems
}

\author{Simone Asci$^{1}$ and Angadh Nanjangud$^{2}$
\thanks{*This work was not supported by any organization}
\thanks{$^{1}$Simone Asci is with the School of Engineering and Materials Science,
        Queen Mary University of London, Mile End Road, London, E1 4NS, United Kingdom
        {\tt\small s.asci@qmul.ac.uk}}%
\thanks{$^{2}$Angadh Nanjangud is with the School of Engineering and Materials Science,
        Queen Mary University of London, Mile End Road, London, E1 4NS, United Kingdom
        {\tt\small a.nanjangud@qmul.ac.uk}}%
}

\begin{document}

\maketitle
\thispagestyle{empty}
\pagestyle{empty}

\begin{abstract}

Collision detection plays a key role in the simulation of interacting rigid bodies. However, owing to its computational complexity current methods typically prioritize either maximizing processing speed or fidelity to real-world behaviors. Fast real-time detection is achieved by simulating collisions with simple geometric shapes whereas incorporating more realistic geometries with multiple points of contact requires considerable computing power which slows down collision detection. In this work, we present a new approach to modeling and simulating collision-inclusive multibody dynamics by leveraging computer algebra system (CAS). This approach offers flexibility in modeling a diverse set of multibody systems applications ranging from human biomechanics to space manipulators with docking interfaces, since the geometric relationships between points and rigid bodies are handled in a generalizable manner. We also analyze the performance of integrating this symbolic modeling approach with collision detection formulated either as a traditional overlap test or as a convex optimization problem. We compare these two collision detection methods in different scenarios and collision resolution using a penalty-based method to simulate dynamics. This work demonstrates an effective simplification in solving collision dynamics problems using a symbolic approach, especially for the algorithm based on convex optimization, which is simpler to implement and, in complex collision scenarios, faster than the overlap test.   

\end{abstract}

\section{INTRODUCTION}

Computer simulation of collision/contact dynamics is a germane research topic of engineering science, particularly within the mechanics \cite{dojo} and robotics \cite{mylapilli} communities.
Current research on computational contact dynamics focuses on the underlying numerical optimisation routines and addresses handling multiple contacts for real-time simulation \cite{multiplepoints}. While important strides have been made here, this research is limited in its employment of predefined models (e.g., 6-DOF manipulators, humanoids, wheeled robots, quadrupeds), that do not generalize to other domains such as spacecraft dynamics and control \cite{pnas}. In this paper, we present a new approach to simulating collision dynamics multibody systems by integrating existing collision detection algorithms with computer symbolic modeling for rigid body dynamics \cite{pendulumconstrained}; compared with traditional approaches, the modeling capability facilitates the application of the algorithm to uncommon and complicated shapes, enabling accurate results in complex scenarios. Further, the symbolic approach is compatible with a variety of contact dynamics models \cite{papadopoulos2021robotic}, which are appropriately formulated to describe the system behavior through symbolic equations of motion. In our work, we exploit an elastic-plastic contact model, commonly used in space manipulator research \cite{contactreview}, \cite{rybus}.

\section{SYMBOLIC APPROACH TO MODEL COLLISION}

\subsection{Symbolic Simulation Framework}

The proposed symbolic modeling approach utilizes SymPy, a widely used computer algebra system (CAS) implemented in Python. We specifically make use of a submodule that derives the symbolic equations of motion (EoMs) of multibody systems \cite{pendulumconstrained}.
The modular design of the framework consists of two parts: the first models the dynamic system and generates the EoMs in symbolic form by means of an automated routine; and the second piece converts symbolic EoMs to their numerical equivalent that can be integrated over time to obtain the evolution of the system.



\subsection{Collision Detection}

Collision detection is defined as the procedure aimed to determine whether two or more objects are overlapping. Specifically, in the context of dynamics simulations, it detects when moving objects are in contact. 
It represents a computational geometry problem with applications in various fields, including computer graphics and video games. Popular algorithms for collision detection are those based on Minkowsky difference, like the Gilbert-Johnson-Keerthi (GJK) algorithm \cite{gilbert1988fast}, \cite{cameron1997enhancing}. Another category of collision detection algorithms are based on the Separating Axis Theorem (SAT) \cite{separatingaxis}, applied in synergy with techniques to approximate the volume occupied by objects, like the Axis-Aligned Bounding Box \cite{aabb}. Such algorithms are employed in several physics engines including Bullet \cite{bullet}, MuJoCo \cite{mujoco}, and Box2D \cite{box2d}, where the trade-off between detection accuracy and computational effort has a major impact on the choice of the algorithm, since collision detection is mainly responsible for simulation slowdown \cite{pnas}. Specifically, the SAT is generally preferred for simple applications, where accuracy plays a secondary role with respect to the availability of computing power. Besides the aforementioned traditional methods, examples of recently proposed methods are the Dynamic Collision Checking which is based on heuristic search \cite{adaptivecollision}, and Fastron which leverages machine learning techniques \cite{fastron}.

\subsection{Collision Resolution}
Collision resolution is defined as the procedure to compute the dynamic behavior of two or more bodies that are in a state of contact with each other. This generally involves resolving the magnitude and direction of contact forces and resulting moments exerted on the interacting bodies to then compute the accelerations, though impulse-based approaches compute only the resulting velocities of the bodies \cite{1995impulse}. The collision force computation relates to dynamics and depends primarily on the materials and mechanical properties of the colliding bodies. Models proposed in the literature develop algebraic models of collision for a pair of objects by running a series of collision experiments and considering the relationship between the pre-impact and post-impact velocities of each body\cite{mote2020collision}. These methods can be classified into two categories: the discrete and continuous models \cite{gilardi2002literature}. The discrete approach, known also as impulse-momentum or complementarity approach, is based on the assumption that the poses of the bodies do not change significantly when short-duration contact occurs; it models bodies as rigid and resolves contact forces using kinematics constraints \cite{1996implicit}. Complementarity methods handle non-smooth events (e.g. collisions and contact interactions) by using impulsive dynamics, unlike continuous methods that simulate body deformation during contact and are affected by issues related to small time step size \cite{jainminimal}. On the other hand, the application of this approach to flexible and multibody systems is complicated. The continuous approach, also referred to as force-based approach or penalty-based method, approximates the local deformation of the contacting bodies using the intersection between their respective geometries, which is then utilised to model the contact force accordingly; it has been widely applied in robotic contact problems, due to its suitability in handling complex geometries \cite{luo2011development}, \cite{wu2017contact}. It includes the bristle-friction \cite{bristle} and the elastic-plastic deformation models \cite{rybus}. The Bristle friction model is based on a linear approximation of the Coulomb friction model. The elastic-plastic model, which is utilized in this paper, calculates the interacting force as the sum of the normal and tangent components to the impact surface, the magnitudes of which depend on the relative velocity of the bodies and the amount of local deformation.


\subsection{Collision response module architecture}

\begin{figure}[htb]
	\centering
        \includegraphics[width=3in]{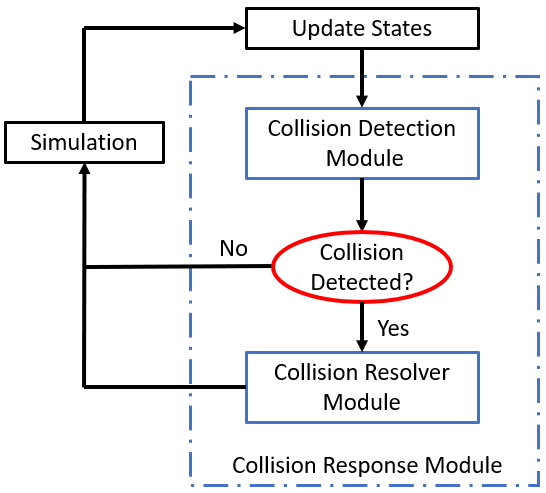}
	\caption{ Architecture of the collision dynamics simulator}
	\label{fig:crespdraft}
\end{figure}

In a typical dynamics simulator, contact interactions are handled by a collision response module (illustrated in Fig. \ref{fig:crespdraft} as the blue dashed box) which takes as input the state of the bodies, i.e. their positions and velocities, and returns a list of forces and moments to update the EoMs for each time step. Its architecture is generally organized into a collision detection module and a collision resolver module; the former detects any collision between the bodies along with any other necessary information needed by the resolver. This generally includes a metric that describes the distance between the bodies, which is referred to as proximity (indicated in this paper with parameter $\phi$, in SI units of $m$), and the minimum distance points (referred to as MDPs in the paper), i.e. the points where the distance between the two bodies is minimum and which therefore are potential future contact points. In case of no overlap, the collision check is terminated and the simulation continues without any update to the EoMs from the collision response module. In this case $\phi$ still contains useful information that can be utilized in the future time steps of the simulation. In case a collision is detected, the collision detection module provides the required information to the resolver which determines all the data necessary for the simulator to resolve the outcome of the collision. Specifically, the interpenetration (indicated with parameter $\rho$, in SI units of $m$) is the metric equivalent to proximity and indicates the depth of penetration between the bodies, while in this case the computed MDPs are actual contact points. MDPs are used to compute the minimum translation vector (MTV) \cite{mtv}, which is defined as the shortest distance along which objects should be moved away to no longer be in a collision state. The direction of the contact force is computed using the MTV, and its magnitude, which depends on the value of $\rho$, is computed according to the specific contact method applied. Once the interacting forces and moments are determined, the resolver module returns them to the simulator which updates the states accordingly.

\section{Collision Detection Methods}
\subsection{Separating Axis Theorem Based Collision Detection}

 \begin{figure}[htb]
	\centering
        \includegraphics[width=3in]{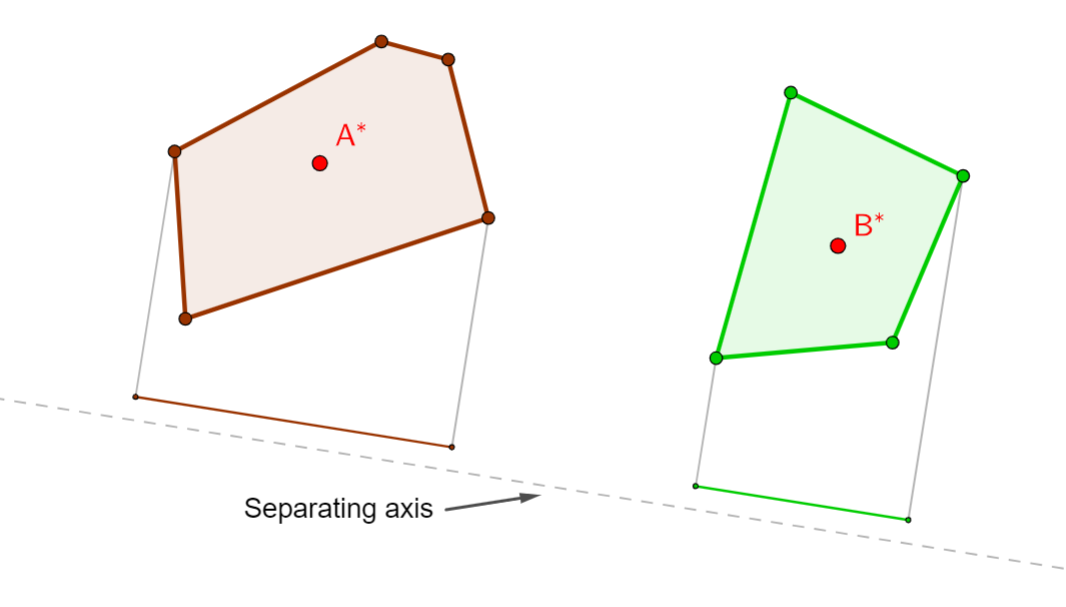}
	\caption{Two convex shapes with their respective projections on a separating axis}
	\label{fig:sat1}
\end{figure}

On a plane, the SAT states that two convex shapes do not intersect if and only if there exists a line for which the projections of the objects onto this line are not overlapping; this line is referred to as a separating axis (Fig. \ref{fig:sat1}).
The same concept extends to the 3D case with a separating plane. The number of lines to be tested depends on the shapes of the objects: two circles require only a single test on the line joining the centers; while for polygons, the normal to each edge of both shapes is a candidate separating axis. Projecting a polygon onto a line requires performing the scalar product between each vertex position and the unit vector lying along the line and storing the range defined by minimum and maximum values; the same operation is repeated for the vertices of the second object. The presence of an overlap can then be determined by a comparison between the two ranges. Therefore the collision detection check consists of a series of scalar products and can provide information also on the amount of interpenetration between shapes, a useful value to perform collision resolution at a later stage. All axes must be tested for overlap to determine intersection, this aspect is responsible for the algorithm's computational complexity and significantly impacts performance, especially when testing polygons with many edges or there are many objects to test. However, at the first axis found where the projections are not overlapping, the algorithm can immediately exit by concluding that the shapes are not in a collision state. Therefore it is possible to speed up the process by appropriately choosing the first axis to test \cite{separatingaxis}, for example, the line joining the centroids of the shapes has the highest probability of detecting the absence of intersection between the shapes. The symbolic approach is particularly advantageous for the SAT implementation because it can handle symbolically the many geometric entities and linear algebra operations required by this algorithm.

\subsection{Convex Optimization Based Collision Detection }

\begin{figure}[htb]
	\centering
        \includegraphics[width=3in]{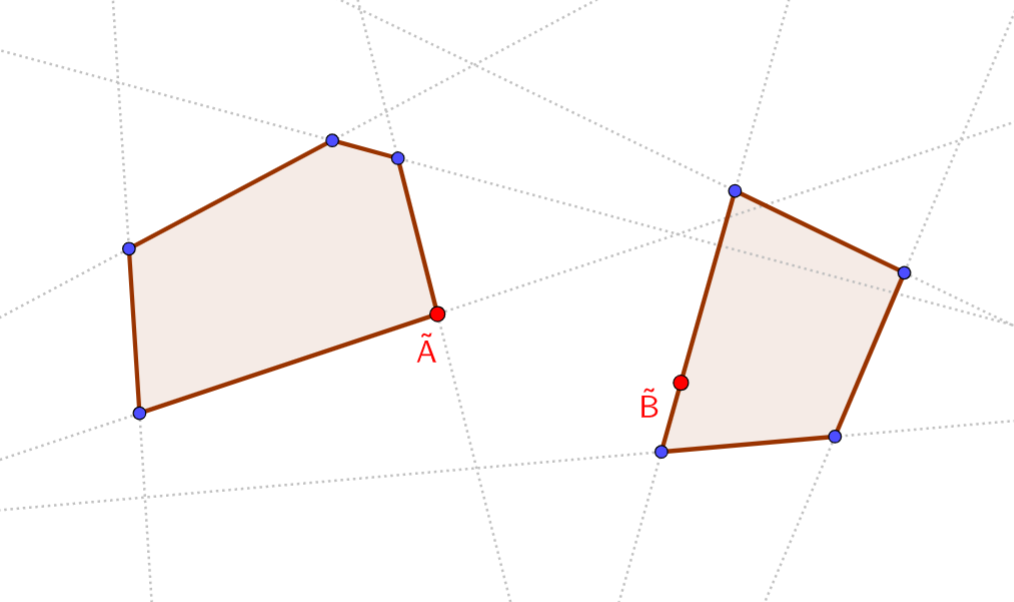}
	\caption{Two convex shapes defined as the intersection of a set of half-planes. The minimum distance points are highlighted in red. }
	\label{fig:convopt}
\end{figure}

Collision detection can also be formulated as an inequality-constrained convex optimization (CO) program that performs a minimization over a continuous objective function in the form \cite{diffpills}:

\begin{mini}|l|
  {x} { \frac{1}{2} \mathbf{{x}^{T}Px} +  \mathbf{{c}^{T}x} }  {}{}
  \addConstraint{ \mathbf{Gx}} {\leq \mathbf{h} }{}
  \label{eq:1}
\end{mini}

\noindent
where $\mathbf{x} \in \mathbb{R}^{n}$ is the independent variable, $\mathbf{P} \in \mathbb{S}_{+}^{n\times n}$ and $\mathbf{c} \in \mathbb{R}^{n}$ are quadratic and linear cost terms; the inequality constraint is described by $\mathbf{G} \in \mathbb{R}^{l\times n}$, and $\mathbf{h} \in \mathbb{R}^{l}$. The returned optimal value for the objective function indicates the proximity, which is positive when no collision is detected and zero if the objects are overlapping. The corresponding optimal values for the independent variables of the problem indicate the coordinates of the MDPs (points $\tilde{A}$ and $\tilde{B}$ in Fig. \ref{fig:convopt}). The MDPs are constrained to be on the objects' boundaries when there is no overlap, or in any point of the intersecting region in case of collision. In this second case, indeed, any pair of overlapping points located in aforementioned region satisfy the constraints of belonging to both objects while having a distance equal to zero, which is an optimal value for the objective function. The MDPs indicate the location where the first contact occurred, consequently providing information to compute the direction of collision force and the amount of interpenetration. Since, in the case of objects intersection, the convex optimization algorithm returns arbitrary values among a set of valid ones for the coordinates of the MDPs, this means that convex optimization-based collision detection is not compatible with penalty-based methods for collision resolution. Under certain circumstances, it is possible to modify the problem to impose the position for the MDPs on the boundaries and correctly compute collision resolution using a method based on interpenetration. However, other methods could generally be used, such as those based on non-penetration constraints, which model the reaction force using Lagrangian multipliers \cite{1996implicit}. In this work, cvxpy \cite{cvxpy} is utilized to solve the convex quadratic program as it offers a variety of solvers based on methods commonly used, including the interior point and the projected gradient methods. We make use of the embedded conic solver (ECOS), an interior-point method for second-order cone programming.


\section{Collision Between Rectangle and Circle}

\begin{figure}[htb]
	\centering
        \includegraphics[width=3in]{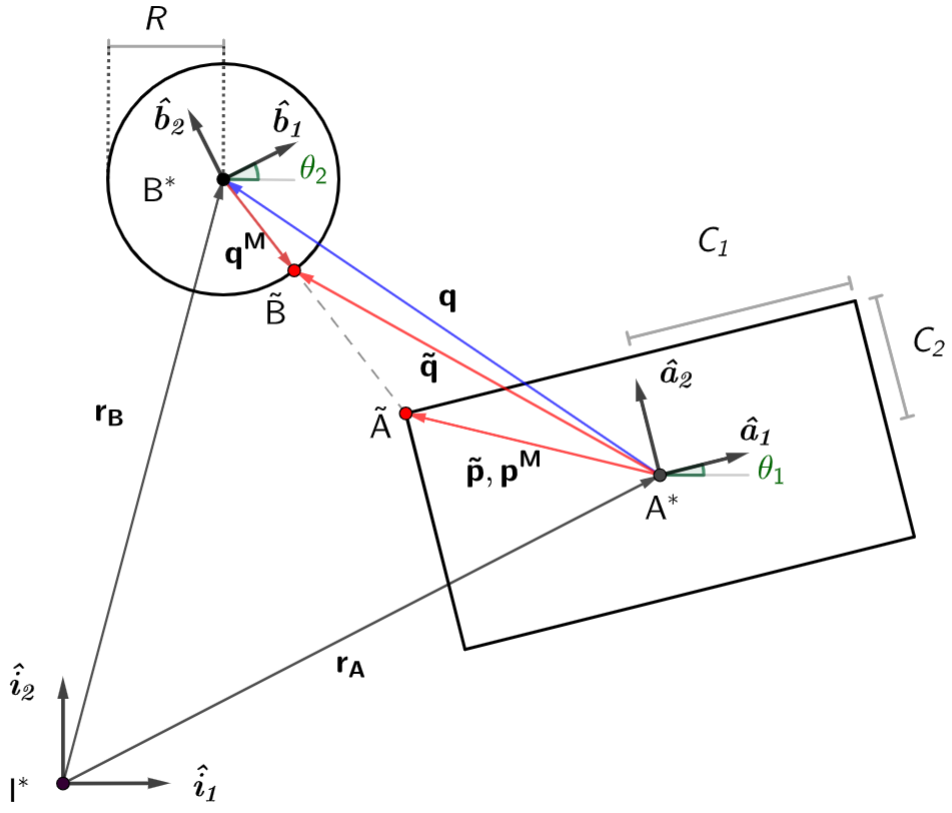}
	\caption{ Geometrical description of a rectangle and a circle as defined by their centroids, orthogonal basis and parameters}
	\label{fig:rect-circle}
\end{figure}

This section describes a rectangle and circle in two-dimensional space (Fig. \ref{fig:rect-circle}). To define the rectangle mathematically, let us consider a point $A^{*}$ in the body-fixed reference frame $\mathbf{A}= (\mathbf{\hat{a}_1, \hat{a}_2, \hat{a}_3)}$ at a distance $\mathbf{r_A} \in \mathbb{R}^{3}$ from $I^{*}$, the origin of an inertial frame $\mathbf{I}= (\mathbf{\hat{i}_1, \hat{i}_2, \hat{i}_3)}$.
Note that $\mathbf{\hat{i}_3}$ and $\mathbf{\hat{a}_3}$ are parallel and out-of-plane. Then any point $y \in \mathbb{R}^{2}$ lying on the plane defined by $\mathbf{\hat{a}_1, \hat{a}_2}$ is within the rectangle if:

\begin{equation}
	  \{  y \mid \mathbf{y} \cdot \mathbf{\hat{a}_1}  \leq C_1 \wedge  \mathbf{y} \cdot \mathbf{\hat{a}_2}  \leq C_2 \} 
\label{eq:2}
\end{equation}

\noindent
where $\mathbf{y}$ is the vector from origin $I^*$ to point $y$, parameters $C_1$ and $C_2$ indicate rectangle dimensions, i.e. half-length and half-width, and $\wedge$ is the logical conjunction operator.

\noindent
Then, consider a circle with center $B^{*}$, of radius $R \in \mathbb{R}$ located at a distance $\mathbf{r_B} \in \mathbb{R}^{3}$ from origin $I^{*}$. Its body-fixed reference frame is defined as $\mathbf{B}= (\mathbf{\hat{b}_1, \hat{b}_2, \hat{b}_3)}$, with $\mathbf{\hat{b}_3}$ parallel to $\mathbf{\hat{i}_3}$. The circle is made up of a set of points on the plane lying on axes $\mathbf{x_I}$ and $\mathbf{y_I}$, such that any point $x \in  \mathbb{R}^{3}$ is within a distance $R$ from the center, as expressed in \eqref{eq:3}:

\begin{equation}
	  \{ x \mid  \left\| \mathbf{x}-\mathbf{r_B} \right\|   \leq R   \wedge  \mathbf{x} \cdot \mathbf{\hat{i}_3} =0  \} 
\label{eq:3}
\end{equation}

\noindent
The following equation shows transformations between frames:

\begin{equation}
\begin{split}
& \mathbf{A} = \mathbf{{}^{I}{R}^{A}} \cdot \mathbf{I} \\
& \mathbf{B} = \mathbf{{}^{I}{R}^{B}} \cdot \mathbf{I} \\
\end{split}
\label{eq:4}
\end{equation}

\noindent
where $\mathbf{{}^{I}{R}^{A}}$ and $\mathbf{{}^{I}{R}^{B}}$ are the rotation matrices relating the inertial frame $\mathbf{I}$ to frames $\mathbf{A}$ and $\mathbf{B}$, respectively. They are defined as follows:

\begin{equation}
\begin{split}
	&\mathbf{{}^{I}{R}^{A}}= 
    \begin{bmatrix}
      \cos \theta_1  & \sin  \theta_1 & 0 \\
    - \sin \theta_1  &  \cos \theta_1   & 0 \\
      0              &      0           & 1
    \end{bmatrix} \\   
    &\mathbf{{}^{I}{R}^{B}}=
    \begin{bmatrix}
      \cos \theta_2  & \sin \theta_2  & 0 \\
    - \sin \theta_2  & \cos \theta_2  & 0  \\
       0              &      0           & 1
    \end{bmatrix} \\
     \label{eq:5}
\end{split}  	
\end{equation}

\noindent
where $\theta_1$ and $\theta_2$ indicate the orientation angles. In the remainder of this section, the vectors are expressed in the frame $\mathbf{A}$, therefore the position of circle center $B^{*}$ is given by $\mathbf{q} =  \mathbf{{}^{I}{R}^{A}}(\mathbf{r_B - r_A})$.

\subsection{Separating Axis Theorem Based Collision Detection}

\begin{figure}[htb]
	\centering
        \includegraphics[width=3in]{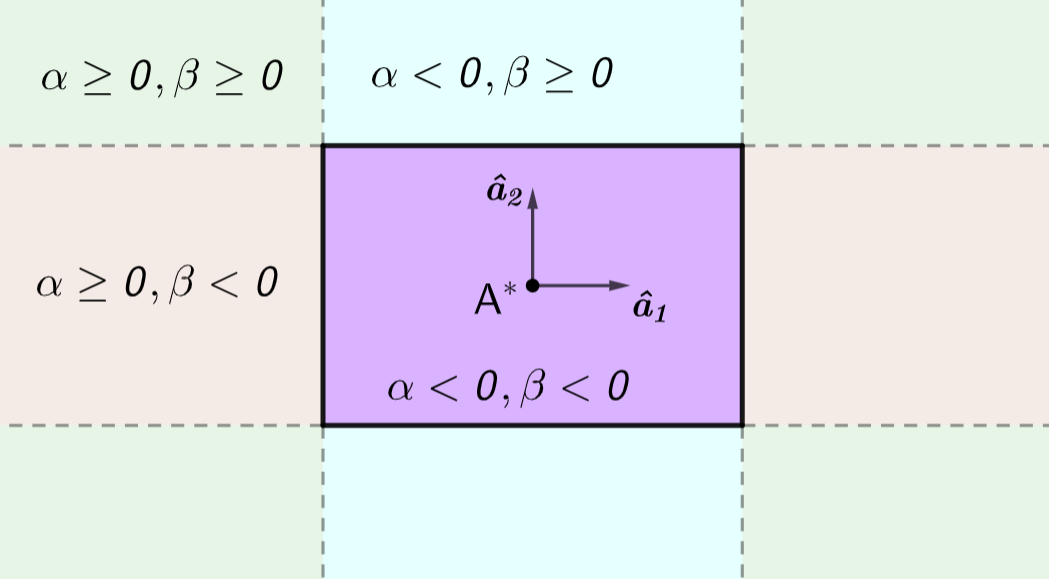}
	\caption{ Rectangle regions}
	\label{fig:rectregions}
\end{figure}

This procedure determines firstly the MDP of the rectangle, which is then used to compute the MDP of the circle and the proximity value $\phi$.
Let us introduce two auxiliary quantities $\alpha$ and $\beta$:

\begin{equation}
\begin{split}
&  \alpha = \lvert  \mathbf{q} \cdot \mathbf{\hat{a}_1}  \rvert - C_1  \\
&   \beta = \lvert  \mathbf{q} \cdot \mathbf{\hat{a}_2}  \rvert - C_2  
\end{split} 
\label{eq:6}
\end{equation}

\noindent
which indicate how far $B^{*}$ is from each rectangle side and in which region of the space it is located; as shown in Fig. \ref{fig:rectregions}, point $B^*$ can be located in:

\begin{itemize}
    \item corner regions and relative vertices, in green 
    \item upper and lower regions and relative edges, in blue
    \item left and right regions and relative edges, in red
    \item region inside the rectangle, in purple
\end{itemize}

\noindent
The MDP of the rectangle is indicated with $\tilde{A}$; it can be any point on its boundary, including a vertex, depending on the circle center location. Its coordinates $x$ and $y$ are defined by \eqref{eq:7}. 

\begin{equation}
\begin{cases}
    [ \text{sgn}(\mathbf{q} {\cdot} \mathbf{\hat{a}_1}) C_1  ,   \text{sgn}(\mathbf{q} {\cdot} \mathbf{\hat{a}_2}) C_2]   \  \text{for $(\alpha, \beta \geq 0)$} \\
        
   [\mathbf{q} {\cdot} \mathbf{\hat{a}_1} ,  \text{sgn}(\mathbf{q} {\cdot} \mathbf{\hat{a}_2}) C_2]  \ \text{for $(\alpha < 0 \wedge \beta \geq 0)$}        \\

   [\text{sgn}(\mathbf{q} {\cdot} \mathbf{\hat{a}_1}) C_1,    \mathbf{q} {\cdot} \mathbf{\hat{a}_2}] \ \text{for $(\alpha \geq 0 \wedge \beta < 0)$}      \\

   [\text{sgn}(\mathbf{q} {\cdot} \mathbf{\hat{a}_1}) C_1,   \mathbf{q} {\cdot} \mathbf{\hat{a}_2} ] \ \text{for $(\alpha,\beta < 0 \wedge  \alpha>\beta )$}      \\
         
    [\mathbf{q} {\cdot} \mathbf{\hat{a}_1} , \text{sgn}(\mathbf{q} {\cdot} \mathbf{\hat{a}_2}) C_2 ]  \ \text{for $(\alpha, \beta < 0 \wedge  \alpha<\beta )$}     \\
          
   [\text{sgn} (\mathbf{q} {\cdot} \mathbf{\hat{a}_1}) C_1 ,   \text{sgn}(\mathbf{q} {\cdot} \mathbf{\hat{a}_2}) C_2 ]  \ \text{for $(\alpha, \beta {<} 0  \wedge  \alpha {=} \beta )$}      \\
\end{cases}
\label{eq:7}
\end{equation}

\noindent
Thus its position vector $\mathbf{\tilde{p}}$ is: 

\begin{equation}
	\mathbf{\tilde{p}} =  x \mathbf{\hat{a}_1} + y \mathbf{\hat{a}_2}  
\label{eq:8}
\end{equation} 

\noindent
Since the collision force is bound to point $\tilde{A}$,
its position with respect to the centroid of the rectangle is given by the vector $\mathbf{p^{M}}$ (Fig. \ref{fig:rect-circle}), which joins point $A^{*}$ to point $\tilde{A}$ and coincides with $\mathbf{\tilde{p}}$. Consequently, the position of the MDP of the circle $\tilde{B}$ with respect to $\mathbf{A}$ frame can be directly computed as: 

\begin{equation}
	\mathbf{\tilde{q}} = \mathbf{q} +  R \frac{\mathbf{\tilde{p}} - \mathbf{q} } { \left\| \mathbf{\tilde{p}} - \mathbf{q}  \right\| }   
\label{eq:9}
\end{equation}

\noindent
The position $\mathbf{q^{M}}$ of the collision force applied to the circle with respect to its center $B^*$ is:
\begin{equation}
	\mathbf{q^{M}} =  R \frac{\mathbf{\tilde{p}} - \mathbf{q} } { \left\| \mathbf{\tilde{p}} - \mathbf{q}  \right\| }   
\label{eq:10}
\end{equation}

\noindent
and $\phi$ is defined as:

\begin{equation}
	\phi =  \left\|  \mathbf{\tilde{p}} - \mathbf{q}  \right\|  - R 
\label{eq:11}
\end{equation}

\noindent
while $\rho$ is given by  \ref{eq:12}.

\begin{equation}
\begin{cases}
\rho = 0  & \text{for $\phi > 0$} \\ 
\rho = \lvert  \phi  \rvert  & \text{for $\phi \leq 0$}   
\end{cases}
\label{eq:12}
\end{equation}

\noindent
The components $n_1,n_2$ of the normal to the surface can be calculated as follows:

\begin{equation}
\begin{cases}
    [ \text{sgn}(\mathbf{q} {\cdot} \mathbf{\hat{a}_1}) C_1 ,  \   \text{sgn}(\mathbf{q} {\cdot} \mathbf{\hat{a}_2}) C_2]   \ \text{for $(\alpha, \beta \geq 0)$} \\
        
    [ 0 , \  \text{sgn}(\mathbf{q} {\cdot} \mathbf{\hat{a}_2}) ]  \ \text{for $(\alpha < 0 \wedge \beta \geq 0)$}        \\

    [\text{sgn}(\mathbf{q} {\cdot} \mathbf{\hat{a}_1}), \  0]   \  \text{for $(\alpha \geq 0 \wedge \beta < 0)$}      \\

   [\text{sgn}(\mathbf{q} {\cdot} \mathbf{\hat{a}_1}), \ 0]  \  \text{for $(\alpha, \beta < 0 \wedge  \alpha>\beta )$}      \\
         
   [ 0, \  \text{sgn}(\mathbf{q} {\cdot} \mathbf{\hat{a}_2})]  \  \text{for $(\alpha, \beta < 0 \wedge  \alpha<\beta )$}     \\
          
    [\text{sgn} (\mathbf{q} {\cdot} \mathbf{\hat{a}_1})  , \  \text{sgn}(\mathbf{q} {\cdot} \mathbf{\hat{a}_2})]  \  \text{for $(\alpha, \beta < 0 \wedge  \alpha=\beta )$}      \\
\end{cases}
\label{eq:13}
\end{equation}

\noindent
and finally the formulas of normal and tangent components of the force:

\begin{equation}
\begin{split}
&   \mathbf{\hat{n}} =  n_1 \mathbf{\hat{a}_1} + n_2 \mathbf{\hat{a}_2} \\
&    \mathbf{\hat{t}} = \mathbf{\hat{a}_3} \times \mathbf{\hat{n}}   
\end{split}
\label{eq:14}
\end{equation}



\subsection{Convex Optimization Based Collision Detection} 

Without loss of generality, all quantities are again expressed in the rectangle reference frame. 
The optimization problem consists of finding two points $\tilde{A}$ and $\tilde{B}$ belonging to the rectangle and circle, respectively, minimizing the square of the distance between them. The algorithm computes correctly the proximity and the MDPs coordinates only when the objects do not intersect because in this case, every point inside the overlapped regions is a candidate to be the MDP, the objective function is always zero and therefore the value of the interpenetration cannot be determined. 
However, a minor modification to the problem formulation makes it possible to calculate $\rho$: by introducing a fictitious circle at the same location as the original one, but with a smaller radius $R^*$ and solving the following convex program for it:

\begin{mini}|l|
  {\mathbf{\tilde{p}, \tilde{q}^*}}  {{\left\|  \mathbf{\tilde{p} - \tilde{q}^*}  \right\| }^2}{}{}
   \addConstraint{  \mathbf{\tilde{p}} \cdot \mathbf{\hat{a}_1}  }{\leq C_1}{}
   \addConstraint{ -\mathbf{\tilde{p}} \cdot \mathbf{\hat{a}_1}  }{\leq C_1}{}
   \addConstraint{  \mathbf{\tilde{p}} \cdot \mathbf{\hat{a}_2}  }{\leq C_2}{}
   \addConstraint{ -\mathbf{\tilde{p}} \cdot \mathbf{\hat{a}_2}  }{\leq C_2}{}
   \addConstraint{{\left\|  \mathbf{q} - \mathbf{\tilde{q}^*}  \right\| }^2}{\leq (R^*)^2}{}
   \label{eq:15}
\end{mini}

\noindent
where $\mathbf{\tilde{q}^*}$ is the position of the MDP belonging to the fictitious circle. Indicating with $b=R-R^*$ the difference between real and fictitious radii, the interpenetration can be recovered from the surrogate proximity $\phi^*$ with \eqref{eq:16}

\begin{equation}
\begin{cases}
\rho = 0  & \text{for $\phi^* > b$} \\ 
\rho = b - \phi^*  & \text{for $ 0 < \phi^* \leq b$}   
\end{cases}
\label{eq:16}
\end{equation}

\noindent
$\phi^*$ is positive also when the original objects are in a collision state, while it becomes zero if the rectangle and fictitious circle come into contact, in which case the interpenetration can no longer be calculated correctly.
Vector $\mathbf{p^M}$ coincides with $\mathbf{\tilde{p}}$, which is returned by the optimization algorithm; while vectors $\mathbf{\tilde{q}}$ and $\mathbf{q^{M}}$ are given by \eqref{eq:9} and \eqref{eq:10} respectively.
Finally the normal and tangent to the rectangle surface are computed as:

\begin{equation}
\begin{split}
&  \mathbf{\hat{n}} = \frac{ \mathbf{\tilde{q}^*} - \mathbf{\tilde{p}} } { \left\|  \mathbf{\tilde{q}^*} - \mathbf{\tilde{p}}  \right\| }   \\
&   \mathbf{\hat{t}} = \mathbf{\hat{a}_3} \times \mathbf{\hat{n}}
\end{split}
\label{eq:17}
\end{equation}




\section{COLLISION RESOLUTION}
In this study, collision resolution is implemented with a penalty-based method. For the sake of simplicity, the reaction force applied at the respective contact point is replaced by an equivalent force-moment system, with the force applied at the center of mass of the body. 
The required information provided by the collision detection algorithm is: 
\begin{itemize}
    \item the amount of interpenetration
    \item normal and tangent to the impact surface
    \item the vector joining each body center of mass to the contact point 
\end{itemize}

\noindent
The magnitude of each force component is computed based on the elastic-plastic approach \cite{rybus}, which defines the interaction between colliding bodies as a spring-damped model. Complete information about material, geometry and velocity of the bodies involved is supposed to be known.
With reference to \eqref{eq:18} the normal force $F_{n}$ is composed of the elastic component, which is proportional to the interpenetration value, and the plastic component, which depends on the relative velocity along the normal direction. The tangent force $F_{t}$ depends on the friction between the surfaces and the relative velocity along the tangent direction.

\begin{equation}
\begin{split}
& F_{N} =   (kc \cdot \rho^{3}) \cdot (1 -cc \cdot v_{N} )  \\
& F_{T} =  -mu \cdot F_{N} \cdot  ( \frac{2}{ 1 + e^{ \frac{-v_{T}}{vs}}  }  -1 )
\end{split}
\label{eq:18}
\end{equation}

\noindent
Where $kc$ and $cc$ are the contact stiffness parameter and damping coefficient respectively, $mu$ is the friction coefficient and $vs$ is a scaling factor, $v_{N}$ and $v_{T}$ are the normal and tangent values of the relative velocity of the bodies. 
The computed force is applied, with opposite signs, to both bodies; while the moment is obtained through a cross-product with the respective force position vector.


\section{RESULTS}


Table \ref{tab:1} compares performances between the symbolic approach and the traditional numeric one in simulating contact dynamics, applying SAT-based method for collision detection and penalty-based method for collision resolution. An additional comparison shows performances of the symbolic approach integrated with CO-based collision detection. Five scenarios were examined by running ten simulations each and calculating the average time, on a computer equipped with an 8-core Intel Core i7-9750H CPU @ 2.60GHz.

\def\tabcolsep{2pt}

\begin{table}[htbp]
    \renewcommand{\arraystretch}{1.3}
    \caption{SIMULATION TIME  }
    \label{tab:1} 
     \centering 
   \begin{tabular}{l|c|c|c } 
    \hline 
    Simulation scenario  & Symbolic SAT   & Numeric SAT  & Symbolic CO   \\
    \hline  
     Bouncing Circle (2D)       &  0.45 \si{s}   &   0.53 \si{s}  &  0.82 \si{s}    \\
     Circle-Circle (2D)         &  1.04 \si{s}   &   0.93 \si{s}  &  0.36 \si{s}   \\
     Rectangle-Circle (2D)      &  1.85 \si{s}   &   1.40 \si{s}  &  1.72 \si{s}   \\
     Rectangle-Rectangle (2D)   &  2.44 \si{s}   &   2.54 \si{s}  &  2.05 \si{s}    \\    
     Sphere-Cuboid (3D)         &  4.01 \si{s}   &   6.83 \si{s}  &  3.09 \si{s}  \\
     \hline
\end{tabular}
\end{table}

\noindent
Regarding 2D scenarios, all methods achieve similar scores, but for the numeric one, this comes at the cost of more modeling effort, which is not reflected in the table. Convex optimization has a clear advantage in the 3D case, where the SAT suffers from a significant increase in the number of tests to be run. Furthermore, the implementation simplicity of CO algorithm scales to more complex cases.

\section{CONCLUSIONS}
This paper presented a novel approach to implementing collision dynamics, through the generation of symbolic equations of motion. The approach was tested in five scenarios, integrating it with the Separating Axis Theorem and convex optimization-based methods for collision detection of multibody systems and with the elastic-plastic approach for collision resolution. We demonstrated that it significantly simplifies the modeling process while retaining the performance advantages of the standard numerical approach; this allows modelers to investigate contact dynamics using a generalizable approach. Further, a comparison between the two tested collision detection methods highlighted that, for 2D cases, the SAT is fast and suitable for applications with simple shapes and few contacts. 
On the other hand, the convex optimization approach is better able to handle complex shapes, although being generally not compatible with penalty-based methods. In applications involving several contacts to check, this research indicates that convex optimization-based collision detection is faster and more accurate than algorithms based on the SAT.



\addtolength{\textheight}{-4cm}   


\bibliographystyle{IEEEtran}
\bibliography{IEEEabrv,references}


\end{document}